\documentclass[10pt,twocolumn,letterpaper]{article}

\usepackage{iccv}
\usepackage{times}
\usepackage{epsfig}
\usepackage{graphicx}
\usepackage{amsmath}
\usepackage{amssymb}

\usepackage{algorithm}
\usepackage{algorithmic}
\usepackage{subcaption}
\usepackage{caption}
\usepackage{url}
\usepackage{array}
\usepackage{colortbl}
\usepackage{booktabs}
\usepackage{multirow}
\usepackage{multicol}
\usepackage{makecell}
\usepackage{hhline}
\usepackage{bbding}
\usepackage{appendix}

\usepackage[breaklinks=true,bookmarks=false]{hyperref}

\iccvfinalcopy % *** Uncomment this line for the final submission

\def\iccvPaperID{****} % *** Enter the ICCV Paper ID here

\ificcvfinal\fi

\begin{document}

%%%%%%%%% TITLE
\title{LVOS: A Benchmark for Long-term Video Object Segmentation}

%% omit the following lines up until the closing ``}''.
%% Additional authors and addresses can be added with ``\and'',
%% just like the second author.
%% To save space, use either the email address or home page, not both
%\and

\author{Lingyi Hong\textsuperscript{1},
		Wenchao Chen\textsuperscript{1},
		Zhongying Liu\textsuperscript{1},
		Wei Zhang\textsuperscript{1,*},\\
		Pinxue Guo\textsuperscript{2},
		Zhaoyu Chen\textsuperscript{2},
		Wenqiang Zhang\textsuperscript{1,2,*}\\
		\textsuperscript{1}Shanghai Key Laboratory of Intelligent Information Processing, \\School of Computer Science, Fudan University\\
		\textsuperscript{2}Academy for Engineering and Technology, Fudan University\\
		{\tt\small lyhong22@m.fudan.edu.cn \{weizh, wqzhang\}@fudan.edu.cn}
	}

\maketitle
\renewcommand{\thefootnote}{*}
\footnotetext[1]{Corresponding author.}
\ificcvfinal\thispagestyle{empty}\fi

%%%%%%%%% ABSTRACT
\begin{abstract}
   Existing video object segmentation (VOS) benchmarks focus on short-term videos which just last about 3-5 seconds and where objects are visible most of the time. These videos are poorly representative of practical applications, and the absence of long-term datasets restricts further investigation of VOS on the application in realistic scenarios. So, in this paper, we present a new benchmark dataset named \textbf{LVOS}, which consists of 220 videos with a total duration of 421 minutes. To the best of our knowledge, LVOS is the first densely annotated long-term VOS dataset. The videos in our LVOS last 1.59 minutes on average, which is 20 times longer than videos in existing VOS datasets. Each video includes various attributes, especially challenges deriving from the wild, such as long-term reappearing and cross-temporal similar objeccts.
		Based on LVOS, we assess existing video object segmentation algorithms and propose a \textbf{D}iverse \textbf{D}ynamic \textbf{Memory} network (\textbf{DDMemory}) that consists of three complementary memory banks to exploit temporal information adequately. The experimental results demonstrate the strength and weaknesses of prior methods, pointing promising directions for further study. 
		Data and code are available at \url{https://lingyihongfd.github.io/lvos.github.io/}.
		
\end{abstract}

%%%%%% table dataset
\begin{table*}
        \setlength{\tabcolsep}{1.6mm} 
		\centering
		%\tiny
        %\scriptsize
        \footnotesize
        \begin{tabular}{l|cccccccccc}
			\toprule
			Dataset     & Videos & \makecell[c]{Mean \\ Frames} & \makecell[c]{Total \\ Frames} & \makecell[c]{Mean \\ Duration}  & \makecell[c]{Total \\ Duration} & \makecell[c]{Frame \\ Rate } & \makecell[c]{Object \\ Classes} & Objects & Annotations & \makecell[c]{Annotations \\ Type}    \\
			\midrule
			FBMS~\cite{ochs2013segmentation}       & 59            & 235          & 13,860         & 0.13          & 7.7           & 30               & 16              & 139     & 1,465      & \textbf{M}             \\
			DAVIS~\cite{perazzi2016benchmark}       & 90          & 69          & 6,298         & 0.04          & 5.17           & 24               & -              & 205     & 13,543      & \textbf{M}      \\
			YouTube-VOS~\cite{xu2018youtube} & \textbf{3,252}       & 27          & 107,181       & 0.06          & 217.2         & 6                & \textbf{78}             & \underline{6,048}   & 133,886     & \textbf{M}         \\
			YouTube-VIS~\cite{yang2019video} & \underline{2,883}          & 28          & 78,000      & 0.06          & 216.7         & 6                & \underline{40}          & 4,883  & $ \sim$131,000    & \textbf{M}            \\	
			OVIS~\cite{qi2022occluded}  & 901          & 90         & $ \sim$68,650      & 0.21       &190.7         & 6                & 25          & \underline{5,223}  &  \underline{$\sim$296,000}     & \textbf{M}            \\	
			UVO~\cite{wang2021unidentified}  & 1,200          & 28          & $ \sim$108,000    & 0.05          & \underline{511}        & 30               & -         & \textbf{14,748}  &  \underline{$\sim$1,327,000}   & \textbf{M}            \\	
			VOT-ST 2021~\cite{kristan2021ninth} &60       & 324        & 19,447     & 0.18          & 10.8        & 30               & -         & 60  & 17,248   & \textbf{M}            \\	
			\midrule
			VOT-LT 2019~\cite{kristan2019seventh}  & 	50       & \textbf{4,305}    & \underline{215,298}         & \textbf{2.39}           & 119              & 30               & -              & 50       & 215,298            & \textbf{B}               \\
			UAV20L~\cite{mueller2016benchmark}     & 	20       & \underline{2,934}   & $\sim$59,000         & \underline{1.63}          & 32.6            & 30               & 5             & 20       & $\sim$59,000             & \textbf{B}               \\
			LaSOT~\cite{fan2019lasot}  & 	\underline{1,400}         & \underline{2,506}    & \textbf{$\sim$3,520,000}       & 1.39           & \textbf{1,950}             & 30               & \underline{70}              & 1,400      & \textbf{$\sim$3,520,000}           & \textbf{B}               \\
			YouTube-VIS 2022 Long~\cite{xu2018youtube}  & 121     & 75         & 9,014      & 0.8         & 100         & 1.5                &-         & -  &  -    & \textbf{N}          \\	
			YouTube-VOS 2022 Long~\cite{yang2019video} & 116        & 67         &7,873     & 0.74          & 87        & 1.5               &-         & 116  &  -    & \textbf{N}           \\	
			Long-time Video~\cite{liang2020video}  & 3           & 2,470        & 7,411         & 1.3           & 4              & 30               & -              & 3       & 60          & \textbf{M} \\
			\textbf{LVOS}        & 220         & 574         & \underline{126,280}       & \underline{1.59}           & \underline{421}            & 6                & 27             & 282     & 156,432      & \textbf{M}   \\
			\bottomrule         
		\end{tabular}
        \caption{Comparison of LVOS with the most popular video segmentation and tracking benchmarks. The top part is existing short-term video datasets and the bottom part is long-term video datasets. Duration denotes the total duration (in minutes) of the annotated videos. Annotations type means the type of groundtruth annotations. \textbf{M} and \textbf{B} denote mask and box annotations. \textbf{N} means that the groundtruth annotations are unavailable. The largest value is in bold, and the second and third largest values are underlined.}
		\label{tab:dataset_static}
\end{table*}

%%%%%%%%% BODY TEXT
\section{Introduction}

\label{sec:intro}
	Given a specific object mask at the first frame, video object segmentation (VOS) aims to highlight target in a video. VOS plays a significant role in video understanding and has many potential downstream applications, such as video editing \cite{oh2018fast}, augmented reality \cite{ngan2011video}, robotics \cite{cohen1999detecting,erdelyi2014adaptive}, self-driving cars \cite{zhang2016instance,ros2015vision,saleh2016kangaroo}. For most practical applications, objects may experience frequent disappearing and videos always last more than 1 minute. It is crucial for VOS model to precisely re-detect and segment target objects in videos of arbitrary length.

However, existing VOS models are specifically designed for short-term situation, which struggle to tackle unforeseen challenges in long-term videos. They are vulnerable to long-term disappearance and error accumulation over time~\cite{voigtlaender2019feelvos,yang2020collaborative,yang2021associating}.~\cite{oh2019video,cheng2021rethinking,hu2021learning,xie2021efficient,seong2020kernelized} may suffer from the poor efficiency and out-of-memory crash due to the ever-expanding memory bank, especially in a long video. However, the lack of the densely annotated long-term VOS datasets restricts the development of VOS in practice. To date, almost all VOS benchmark datasets, such as DAVIS~\cite{perazzi2016benchmark} and YouTube-VOS~\cite{xu2018youtube}, just focus on short-term videos, which are a poor reflection of practitioners’ demands. The average video length is less than 6 seconds and target objects are always visible, while the average duration is much more longer (\textit{i}.\textit{e}.,1-2 minutes) and target objects disappear and reappear frequently in real-world scenarios.

	To this end, we propose the first \textbf{long-term} video object segmentation benchmark dataset, named \textbf{L}ong-term \textbf{V}ideo \textbf{O}bject \textbf{S}egmentation (\textbf{LVOS}). LVOS contains 220 videos with an average duration of 1.59 minutes. The emphasized properties of LVOS are summarised as follows. (1) \textbf{Long-term}. 
	Videos in LVOS last 1.59 minutes on average (\textit{vs} 6 seconds in short-term videos), which is much closer to real applications (Table \ref{tab:dataset_static}). These videos cover multiple challenges, especially attributes specific in long-term videos such as frequent reappearance and long-term similar object confusion. Figure \ref{fig:data_overview} shows some  sample videos. 
	(2) \textbf{Dense and high-quality annotations}. All frames in LVOS are manually and precisely annotated at 6 FPS. To annotate the target object accurately and efficiently, we build a semi-automatic annotation pipeline. 
	There are 156K annotated objects in LVOS, about 18\% times more annotations than the largest VOS dataset~\cite{xu2018youtube}. 
	(3) \textbf{Comprehensive labeling.} Videos in LVOS feature 27 categories to represent the daily scenarios. Among the 27 categories, there are 7 unseen categories to better assess the generalization ability of models.

	Extensive experiments on LVOS are conducted to assess existing VOS models. 
	To capture the different temporal context in long-term videos adequately, we propose \textbf{D}iverse \textbf{D}ynamic \textbf{Memory} (\textbf{DDMemory}). DDMemory consists of three complementary memory banks: reference memory, global memory, and local memory to encode historical information into fixed-size features. Due to the diverse and dynamic memory mechanism, DDMemory can handle videos of any length with constant memory cost and high efficiency. Oracle experiment demonstrates that error accumulation and complex motion are the primary cause for the unsatisfactory performance.

	Our contributions are summarized as follows: (1) We construct  a new long-term, densely and high-quality annotated, and comprehensively labeled VOS dataset named LVOS with 220  videos whose average duration is 1.59 minutes. (2) We propose the DDMemory to handle long-term videos better. (3) We assess existing VOS models and DDMemory on LVOS and analyze the cause of errors to discover cues for the development of robust VOS methods.
	%%%%% related work
	\section{Related work}
	\textbf{Semi-supervised Video Object Segmentation.} The key to semi-supervised VOS lies in the construction and utilization of feature memory. \cite{caelles2017one,maninis2018video,voigtlaender2017online,xiao2018monet,robinson2020learning, meinhardt2020make, park2021learning, bhat2020learning} employ online learning approaches to finetune pretrained networks at test time on the first frame and groundtruth, which require a large amount of time. \cite{hu2018videomatch,chen2018blazingly,shin2017pixel,bao2018cnn} employ the manually annotated first frame to guide the segmentation of the rest frames, while \cite{perazzi2017learning,zhang2019fast,cheng2018fast,jang2017online,chen2020state,oh2018fast,xu2019spatiotemporal,hu2018motion,ventura2019rvos,hu2017maskrnn,li2018video,wang2019fast,khoreva2017lucid} use the already segmented previous frame as a reference to propagate mask frame-to-frame. \cite{voigtlaender2019feelvos,yang2020collaborative,yang2018efficient,wang2019ranet,johnander2019generative} combine the first frame and previous frame as feature memory, and the temporal context is limited. To address the limitation, \cite{oh2019video,9665289,cheng2021modular,seong2020kernelized,hu2021learning,xie2021efficient,wang2021swiftnet,lu2020video,cheng2021rethinking,seong2021hierarchical,mao2021joint,yang2021associating,park2022per,lin2022swem} develop a feature memory bank to store all historical frames, while the ever-expanding memory bank may encounter an out-of-memory crash when handling long-term videos.

	The adaptive feature bank is developed in~\cite{liang2020video} to dynamically manage key features of objects by using exponential moving averages.~\cite{li2020fast} introduces a global context module to summarize target information.~\cite{li2022recurrent} proposes a recurrent dynamic embedding (RDE) to construct the memory bank of fixed size in a recurrent manner. Xmem~\cite{cheng2022xmem} develops three kinds of memory banks and connects them to segment the current frame. By compressing the memory bank, these methods achieve constant memory cost, but they still struggle with losing track after a long period of disappearing in long-term videos. Thus, we propose DDMemory to  efficiently exploit the temporal context and maintain the fixed-sized memory cost, which is robust to various challenges in long-term videos.

    \textbf{Short-term Video Object Segmentation Dataset.} The existing video object segmentation benchmark datasets are all short-term video datasets.
	FBMS~\cite{ochs2013segmentation} has 59 sequences with 13,860 frames in total and is divided into 29 and 30 videos as training and evaluation set, respectively. DAVIS 2017~\cite{pont20172017} is a popular benchmark dataset with 60 and 30 videos for train and validation sets. There are 6,298 frames in total. DAVIS 2017 provides pixel-level and high-quality annotations for each frame. YouTube-VOS~\cite{xu2018youtube}, as a large-scale dataset, has 3,252 sequences with precise annotations at 6 FPS.  YouTube-VOS includes 78 diverse categories. All these benchmarks are short-term video datasets, where the average duration of videos is about 3-6 seconds. 
Although some VOS methods~\cite{liang2020video,li2020fast,li2022recurrent,cheng2022xmem} claim to scale well to long-term videos, they do not conduct quantitative experiments on a long-term VOS benchmark because of the lack of such a dataset. Videos in LVOS are long-term with an average duration of about 1.59 minutes, which is relevant to scenarios for actual applications.

    \textbf{Long-term Tracking Dataset.}
	There are several benchmark datasets specific to long-term tracking. UAV20L \cite{mueller2016benchmark} is a small scale dataset with only 20 long videos. OxUvA \cite{valmadre2018long} consists of 366 sequences, but each video is sparse-annotated every 30 frames. LaSOT~\cite{fan2019lasot} is the first large-scale and densely annotated long-term tracking dataset, which provides 1,400 videos totaling 3.52M frames. The average length of sequences in LaSOT is 2,512 frames at 30 FPS. Each frame is manually annotated with a bounding box. These long-term tracking datasets demonstrate the significance of long-term tasks. However, these datasets only provide box-level annotations, and pixel-level annotations are unavailable, which is more crucial for fine-grained study. Long-time Video~\cite{liang2020video} is a dataset of 3 long videos with 2,470 frames on average per video, where only 20 frames are uniformly annotated for each video. 
Note that YouTube-VOS 2022 Long and YouTube-VIS 2022 Long proposed at the CVPR 2022 workshop also include long-term videos, while no groundtruth is available. There lacks a comprehensive long-term VOS dataset which is with training data and available all the time. 
LVOS focuses on long-term video object segmentation with 220 videos in total including both training, valid, and test sets. Each frame in LVOS is manually and precisely annotated. We propose LVOS to promote the development of robust VOS models and provide a more suitable evaluation benchmark in practical application.

\begin{figure*}[htbp]
		\centering
		\includegraphics[width=1.0\linewidth]{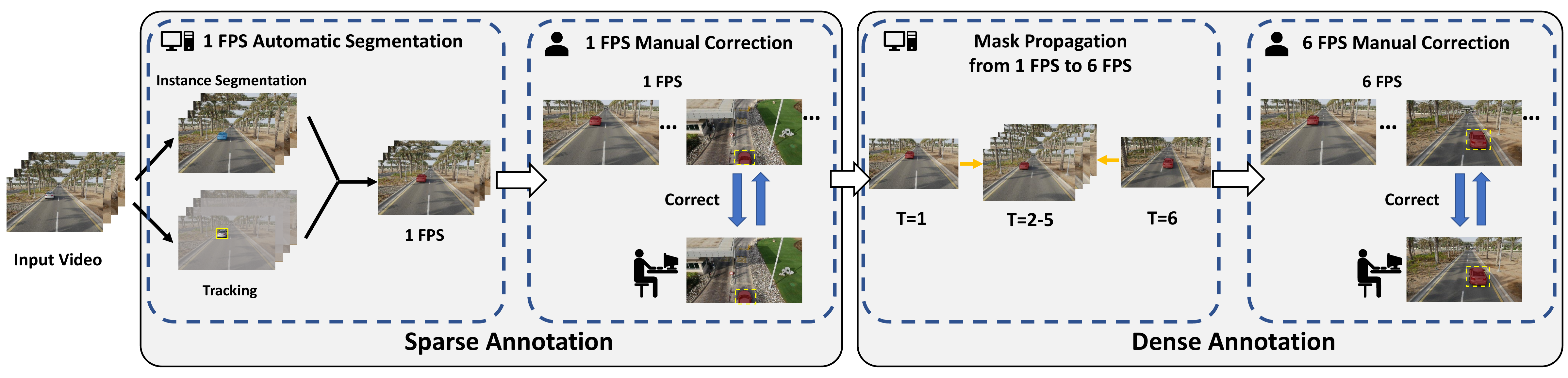}
		
		\caption{Annotation Pipeline, including four steps. Step 1: 1 FPS Automatic Segmentation. We utilize instance segmentation \cite{ke2022mask} and tracking \cite{cui2022mixformer} models to get the mask of target object at 1 FPS automatically. Step 2: 1 FPS Manual Correction. We refine masks obtained in Step 1 manually. Step 3: Mask Propagation from 1 FPS to 6 FPS. We propagate masks from 1 FPS to 6 FPS by using a VOS model\cite{yang2021associating}. Step 4: 6 FPS Manual Correction. We correct the masks obtained in Step 3 manually. }
		\label{fig:anno_pipeline}
\end{figure*}

\section{LVOS: Long-term Video Object Segmentation Benchmark Dataset}
	\subsection{Dataset Construction}
	\textbf{Dataset Design.} To make up for the lack of a dedicated dataset, LVOS aims to provide the community with a novel and dedicated VOS dataset for training and evaluating robust VOS models. We adhere to the three principles listed below to construct LVOS.
	
	1) \textbf{Long-term VOS.} Compared with current VOS  datasets \cite{perazzi2016benchmark,xu2018youtube} where the average length of each video is only 3-6 seconds, we ensure videos in LVOS last about 1.59 minutes (\textit{i}.\textit{e}., 574 frames at 6 FPS), about 20 times longer than short-term videos, which is much closer to the real application.

	2) \textbf{Dense and high-quality annotation.} The time-consuming mask annotation processing severely constrains the duration and scale of current VOS datasets. High-quality and densely annotated masks are essential for training robust VOS models and assessing their performance in practical applications. So, all frames in LVOS are manually and precisely annotated by leveraging the semi-automatic annotation pipeline proposed in Sec \ref{sec:pipeline}.
	
	3) \textbf{Comprehensive labeling.} We design a set of categories that are relevant to daily life and have 5 parent classes and 27 subclasses. It is worth noting that the 27 categories are not limited to COCO dataset~\cite{lin2014microsoft} and also include some categories not present in the COCO dataset, such as frisbee. Among the 27 categories, there are 7 unseen categories to better assess the generalization ability of models.

	\textbf{Data Collection.}
	To construct LVOS, we carefully select a set of categories comprising 5 parent classes and 27 subclasses from the videos in existing long-term tracking datasets such as VOT-LT~\cite{kristan2019seventh} and LaSOT~\cite{fan2019lasot}. These datasets, containing more than 1,800 videos in total, have been customized for long-term tracking are similar to VOS in terms of tracking tasks. As such, the videos in VOT-LT and LaSOT are suitable for the long-term VOS task. After selecting  a set of object categories, we screen about 600 videos with a resolution of 720P as candidate videos. 220 videos are selected to make up for LVOS after a comprehensive consideration of video quality. Because videos in VOT-LT and LaSOT have been processed specially for tracking task, such as removing irrelevant content, we did not apply any treatment to these videos. VOT-LT and LaSOT are single-object datasets, where only one target is annotated in a video, while our LVOS is multiple-object, where there may be several target objects in a video. For target selection, we may either follow the target object in VOT-LT and LaSOT, or select different objects as targets.
	
	\subsection{Semi-Automatic Annotation Pipeline}
	\label{sec:pipeline}
	The exhaustion of the mask annotation process limits the scale of VOS datasets to a large extent. We propose a novel semi-automatic annotation pipeline to annotate frames efficiently. Concretely, the pipeline can be divided into four steps, as shown in Figure \ref{fig:anno_pipeline}.
	
	\textbf{Step 1: 1 FPS Automatic Segmentation.} 
	Firstly, transfiner \cite{ke2022mask} is adopted to generate the pixel-wise segmentation of each object in the frames at 1 FPS. Then we manually mark the bounding box of the target objects when they first appeared and utilize MixFormer \cite{cui2022mixformer} to propagate the box from the first frame to all subsequent frames. Based on the pixel-wise segmentation and the bounding box of each frame, we obtain the masks of target objects at 1 FPS.
	
	\textbf{Step 2: 1 FPS Manual Correction.} Tracking errors, segmentation defects, and other prediction mistakes may lead to inaccuracy or the absence of the target object mask in some frames. Thus, we use EISeg \cite{hao2021edgeflow} (An Efficient Interactive Segmentation Tool based on PaddlePaddle \cite{ma2019paddlepaddle}) to refine masks. On average, about 30\% frames need to be corrected.
	
	\textbf{Step 3: Mask Propagation.} By using a VOS model (\textit{i}.\textit{e}., AOT \cite{yang2021associating}) to propagate the annotation masks at the frame rate of 1 FPS obtained in Step 2 to their adjacent unlabeled frames, we extend the masks from 1 FPS to 6 FPS automatically. 
	
	\textbf{Step 4: 6 FPS Manual Correction.} Because of flaws in masks segmented by VOS model, we correct every frame artificially until the results are satisfactory. In this step, about 40\% of frames require further refinement. 
	
	\textbf{Time and Quality Analysis.} To examine the annotation quality, we randomly choose 100 videos from HQYouTube-VIS~\cite{ke2022mask} training set and relabel them using our semi-automatic annotation pipeline. Then we compare the results with the groundtruth, and the average IoU score is 0.93. The score shows that the annotation results obtained by our pipeline are largely consistent with groundtruth and also proves the validity of our pipeline. Moreover, we ask annotators to record the total time overheads. It takes 60 minutes for one annotator to label an entire long-term video (500 frames at 6 FPS) on average by utilizing our pipeline, while a skilled annotator spends 1500 minutes labeling the same video (3 minutes for one frame). The pipeline significantly reduces the labeling cost when ensuring annotation quality.

%%%% attribute figure
	\begin{figure*}[htbp]
		\centering
		\begin{subfigure}{0.33\linewidth}
			\includegraphics[width=1.0\linewidth]{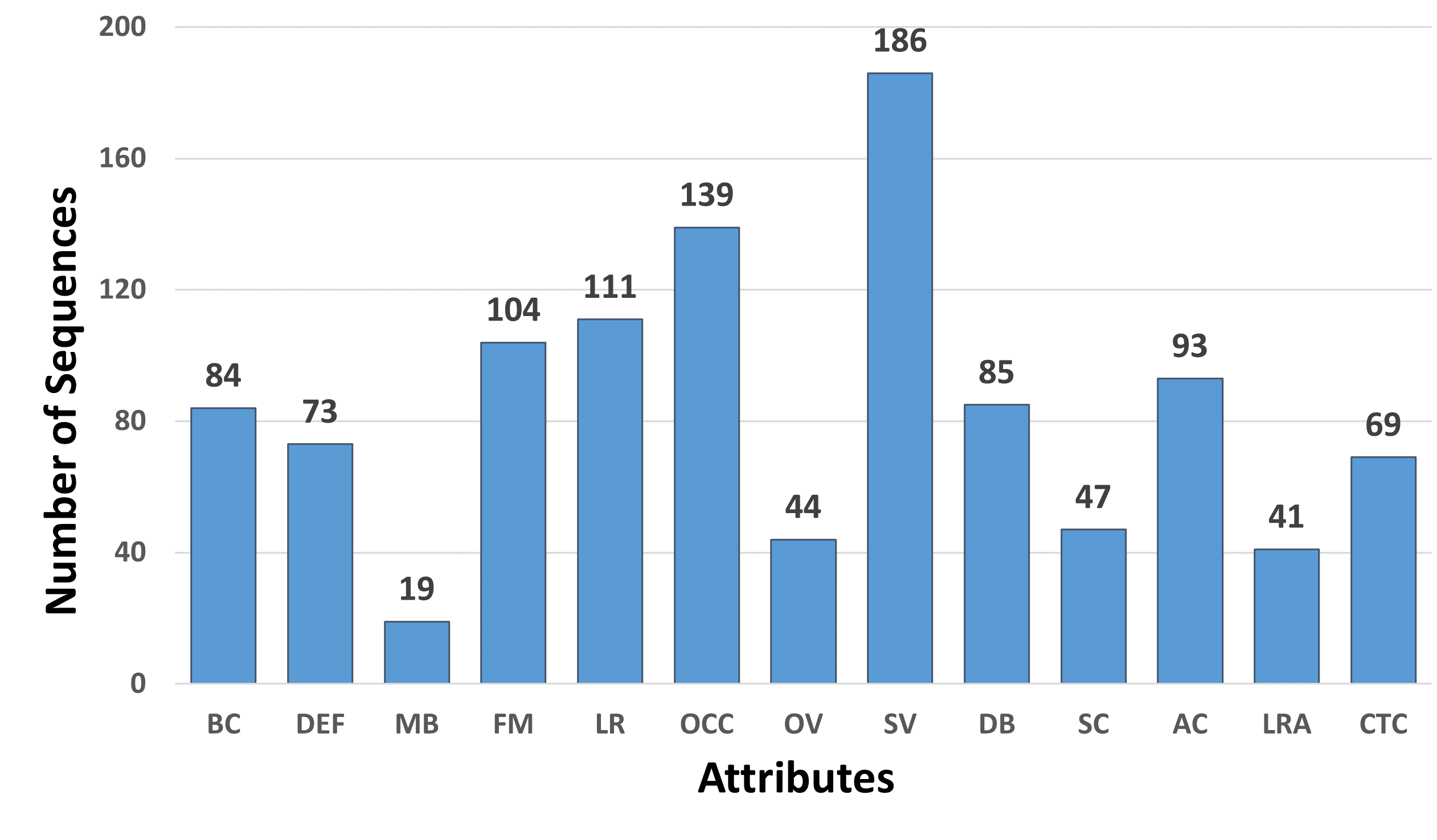}
			\caption{Attributes distribution of each sequence in LVOS.}
			\label{fig:data_attribute_table}
		\end{subfigure}
		\hfill
		\begin{subfigure}{0.28\linewidth}
			\includegraphics[width=1.0\linewidth]{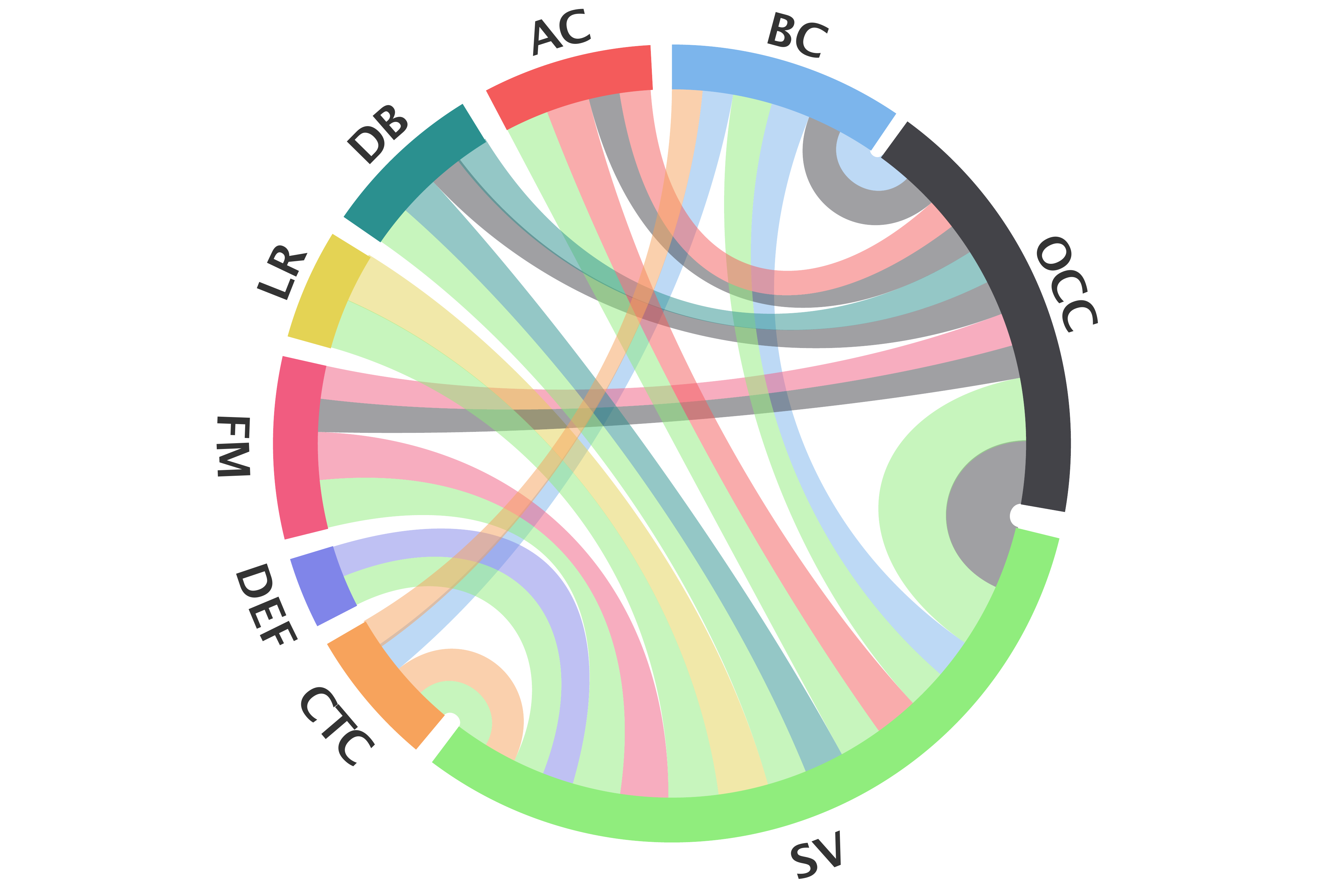}
			\caption{Main Mutual dependencies among attributes.}
			\label{fig:data_attribute_chart}
		\end{subfigure}
		\hfill
		\begin{subfigure}{0.33\linewidth}
			\includegraphics[width=1.0\linewidth]{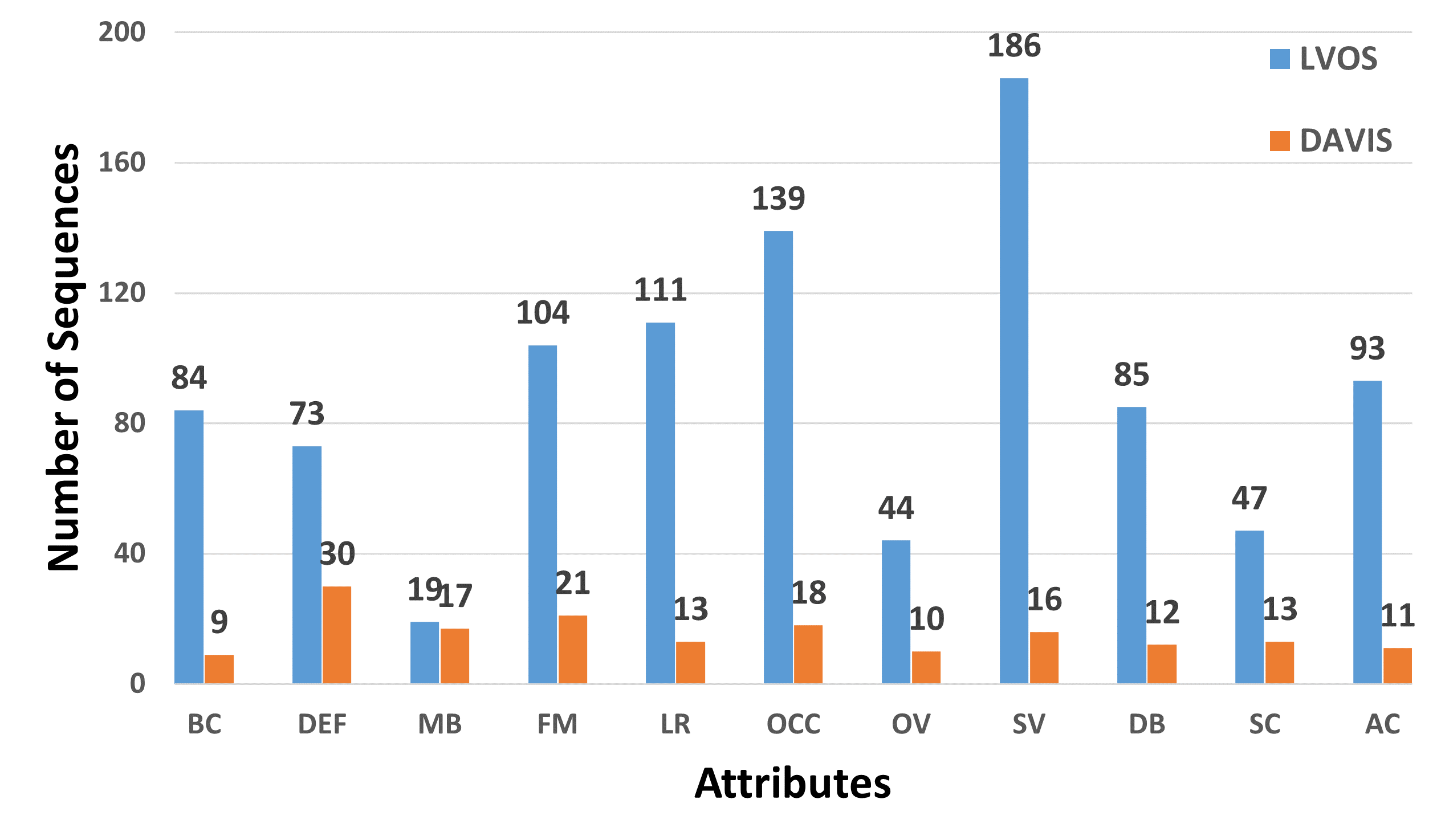}
			\caption{Distribution comparison with DAVIS 2017~\cite{perazzi2017learning} .}
			\label{fig:data_attribute_compare}
		\end{subfigure}
		\caption{ Attributes distribution in LVOS and comparison with DAVIS2017~\cite{perazzi2017learning}. In sub-figure (b), the link indicates the high likelihood that more than one attributes will appear in a sequence. Best viewed in color.}
		\label{fig:data_attribute}
	\end{figure*}

    \textbf{Discussion.} Similar semi-automatic annotation pipelines have also been proposed in UVO~\cite{wang2021unidentified} and EPIC-VISOR~\cite{darkhalil2022epic}, which can be seperated into two parts: manual annotation of videos sparsely and propagating masks. The mask propagation part is similar to the Step 3 and Step 4 in our pipeline. For sparsely annotating video, pipelines in UVO and EPIC-VISOR require manual annotation, while our pipeline also adopts models, which is faster. To obtain the pixel-wise segmentation of each object in Step 1, we employ Transfiner trained on COCO. Classes in LVOS are not restricted to classes in COCO or LVIS~\cite{gupta2019lvis}. Transfiner can detect and segment the  segment objects whose classes are not among the 80 categories in COCO, although the category classification is incoerrect. Because VOS is class-agnostic, we simply ignore the category and utlize the mask segmentaion, where classification errors have no influence on the annotation pipeline.

	%%%% instance numbers
	\begin{figure}[htbp]
		\centering
		\includegraphics[width=1.0\linewidth]{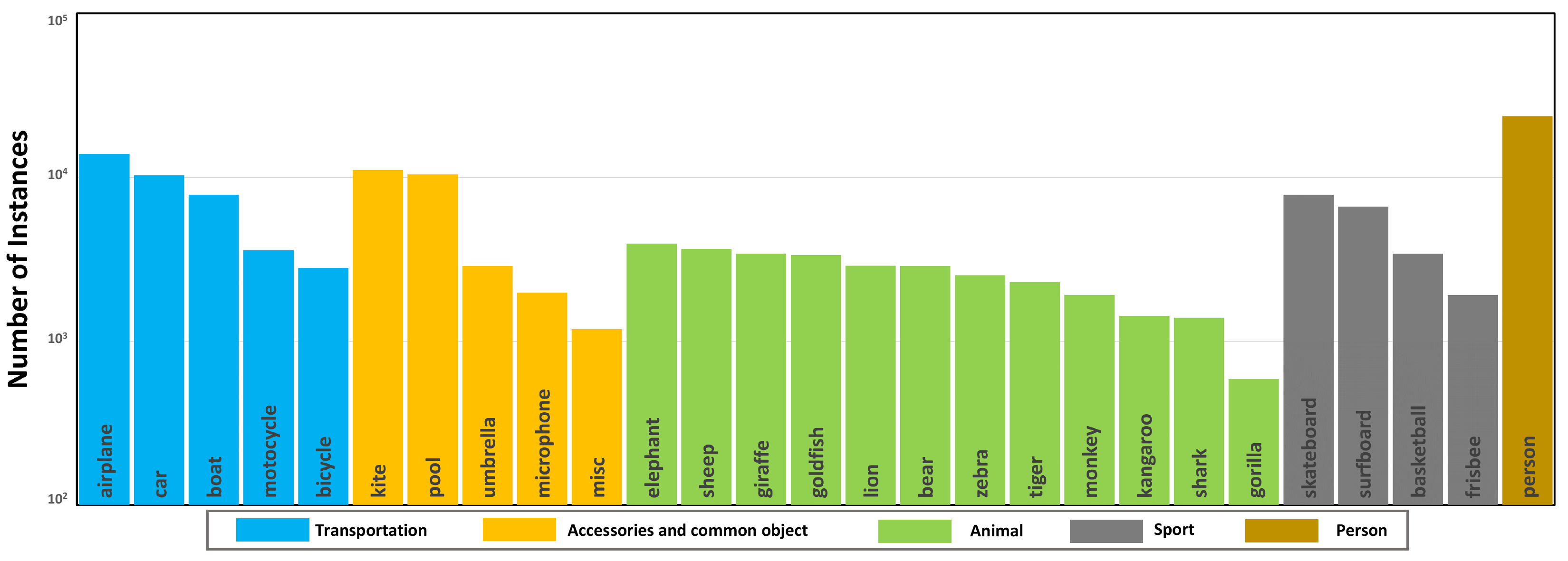}
		\caption{The histogram of instance masks for five parent classes and sub-classes. Objects are sorted by frequency. The entire category set roughly covers diverse objects and motions that occur in everyday scenarios. }
		\label{fig:instance_nums}
	\end{figure}

%%%%%% table challenge
	\begin{table}[htbp]
		\scriptsize 
		\centering
		\begin{tabular}{ll}
			\toprule
			Attribute & Definition  \\
			\midrule
			BC        &  \textit{Background Clutter.} The appearances of background and target  \\
			& object are similar.  \\
			DEF       & \textit{Deformation.} Target appearance deform complexly.  \\
			MB        & \textit{Motion Blur.} Boundaries of target object is blurred because of   \\ 
			&  camera or object fast motion.\\
			FM        & \textit{Fast Motion.} The per-frame motion of target is larger than 20 \\
			&    pixels, computed as the centroids Euclidean distance.\\
			LR        & \textit{Low Resolution.} The average ratio between target box area and  \\ 
			&    image area is smaller than 0.1 . \\
			OCC       & \textit{Occlusion.} The target is partially or fully occluded in the video. \\
			OV        & \textit{Out-of-view} The target leaves the video frame completely.\\
			SV        & \textit{Scale Variation} The ratio of any pair of bounding-box is outside \\
			&  of range {[}0.5,2.0{]}.                  \\
			%\midrule
			DB        & \textit{Dynamic Background} Background regions undergos deformation. \\
			SC        & \textit{Shape Complexity} Boundaries of target object is complex. \\
			AC        & \textit{Appearance Change} Significant appearance change, due to rota-  \\      
			& tions and illumination changes . \\
			\midrule
			LRA       & \textit{Long-term Reappearance} Target object reappears after disappear- \\
			&  ing for at least 100 frames. \\
			CTC       & \textit{Cross-temporal Confusion} There are multiple different objects that   \\
			&   are similar to targect object but do not appear at the same time.\\
			\bottomrule            
		\end{tabular}
		\caption{Definitions of video attributes in LVOS. We extend and modify the short-term video challenges defined in \cite{perazzi2016benchmark} (top), which is exanded with a complementary set of  long-term video attributes (bottom).}
		\label{tab:challenge}
	\end{table}

	\subsection{Dataset Statics}
	\textbf{Video-level Statics.}
	The video-level information of LVOS is shown in Table \ref{tab:dataset_static}. We collect 220 videos in LVOS, whose average duration is 1.59 minutes with 574 frames on average at 6 FPS (\textit{vs} 3-6 seconds in short-term dataset). There are 126,280 frames and 156,432 annotations in total, which is larger than the sum of the other data sets \cite{perazzi2016benchmark,xu2018youtube,liang2020video,ochs2013segmentation}. Videos are categorized into 5 parent classes and 27 sub-classes. The detail and distribution of instance masks can be seen in Figure \ref{fig:instance_nums}. Notably, there are 7 categories which are not present in the training set. We sample frames with a frame rate of 6 FPS. By keeping the distribution of subsets and video length, videos are divided into 120 training, 50 validation, and 50 testing. Annotations of the training and validation sets are publicly released for the development of VOS methods, while annotations of the testing set are kept private for competition use.

	%%%% model
	\begin{figure*}[htbp]
		\centering
		\includegraphics[width=\linewidth]{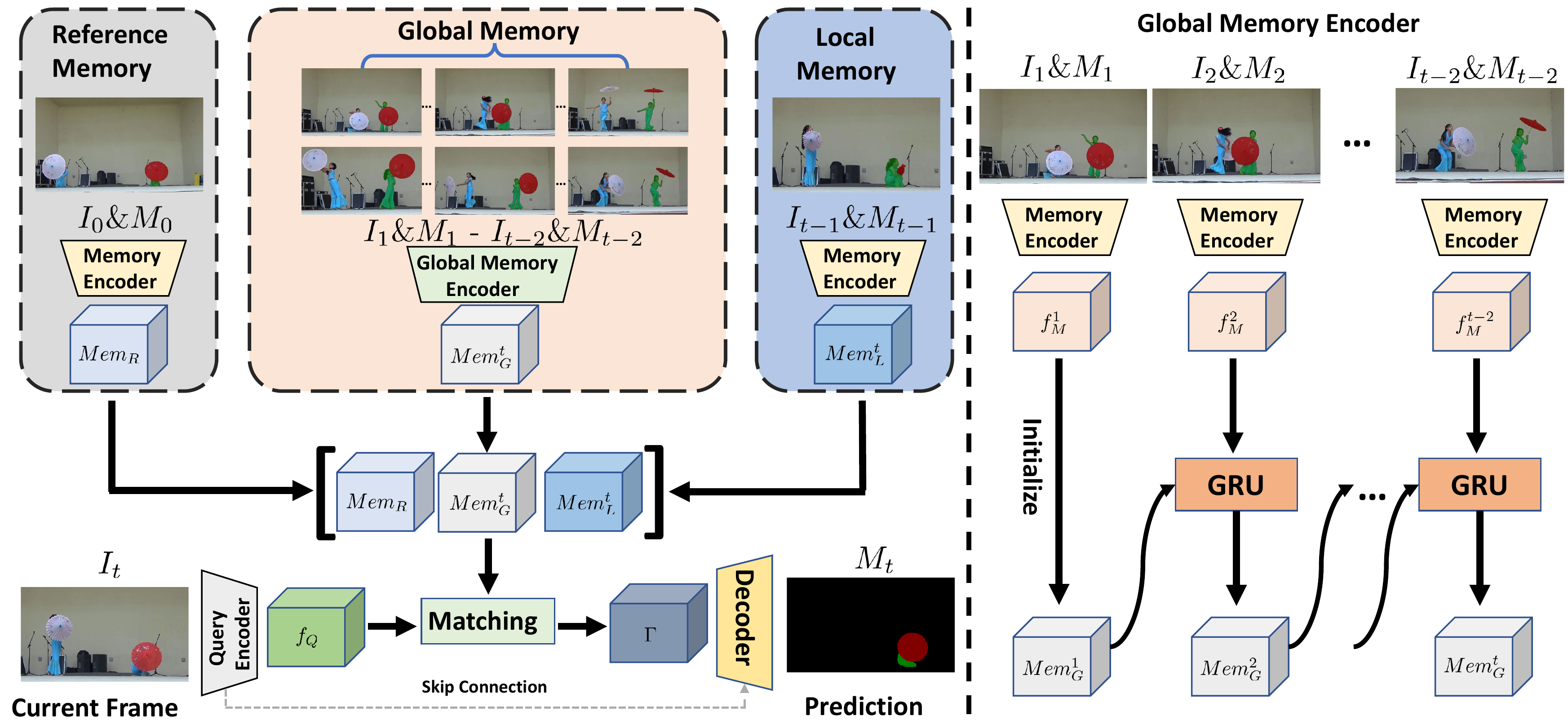}
		
		\caption{Model Overview. DDMemory consists of three memory banks: reference memory, global memory, and local memory. The global memory encoder is responsible for encoding the historical information into a fixed-size feature. }
		\label{fig:model}
	\end{figure*}

\textbf{Attributes.}
	For a further and comprehensive analysis of VOS approaches, it is of great significance to identify video attributes. We label each sequence with 13 challenges, which are defined in Table \ref{tab:challenge}. These attributes include short-term video challenges, which are extended from DAVIS \cite{perazzi2016benchmark}, and are expanded with a complementary set of challenges specific to long-term videos. It is important to note that these attributes are not exclusive, and a video can contain multiple challenges. The distribution of each video and the main mutual dependencies are shown in Figure \ref{fig:data_attribute_table} and \ref{fig:data_attribute_chart}. Scale variation (SV), occlusion (OCC), low resolution (LR), and fast motion (FM) are the most common challenges in LVOS. The comparison of attributes distribution between LVOS and DAVIS is demonstrated in Figure \ref{fig:data_attribute_compare}. We observe the difference in challenges between short-term and long-term videos. Because of the longer length of videos, the object motion and background changes are much more complex and varied, which is not obvious in short-term videos. The variation in the distribution of attributes places differences and higher demanding requirements on the design of VOS models.

% method
	\section{Method}
	Although many VOS methods \cite{liang2020video,li2020fast,li2022recurrent,cheng2022xmem,lin2022swem,wang2021swiftnet} attempt to compress memory bank to  achieve the trade-off of efficiency and accuracy, these models still struggle with the loss of critical temporal information in long videos. We propose a novel VOS method, \textbf{D}iverse \textbf{D}ynamic \textbf{Memory} (\textbf{DDMemory}), specifically designed for long-term VOS task, which includes diverse memory banks with constant size: \textit{reference memory}, \textit{global memory}, and \textit{local memory}. 
	Diverse memory banks can compress the global temporal memory into three memory features with rich temporal information and maintain low GPU memory usage while achieving high performance.

	\subsection{Method Overview}
	Let ${I_{1}}$ and ${M_{1}}$ denote the first frame and its groundtruth mask, and ${I_{t}}$ denotes the current frame whose segmentation mask ${M_{t}}$ is to be predicted. $\{{{I_{n}},{M_{n}}}\}^{t-1}_{n=2}$ denote the intermediate frames from the second frame to the previous frame, along with their estimated masks. Following \cite{yang2021associating}, for the query frame ${I_{t}}$ of size $H \times W$, query encoder takes the image as input to extract visual features $f_{Q} \in \mathbb{R}^{C \times \frac{H}{16} \times \frac{W}{16}} $, where $C$ is the channel dimension. Each intermediate frame and corresponding mask are fed into memory encoder to obtain memory feature $f_{M}^{i} \in \mathbb{R}^{C \times \frac{H}{16} \times \frac{W}{16}}  (i=0,1,\cdots,n-1)$.  The memory feature of first and previous frame, $f_{M}^{0}$ and $f_{M}^{t-1}$, act as the reference memory $Mem_{R}$ and local memory $Mem_{L}^{t}$, respectively. For $\{{f_{M}^{i}}\}^{t-2}_{i=2}$,  we feed them into global memory encoder (Sec \ref{sec:longencoder}) to generate global memory feature $Mem_{G}^{t} \in \mathbb{R}^{C \times \frac{H}{16} \times \frac{W}{16}}$. The three memory features ($Mem_{R}$, $Mem_{G}$, $Mem_{L}$) are fed into the matching module together with the query feature $f^{Q}$ to get the matching output $\Gamma$. Finally, $\Gamma$ and the low-level features from the decoder are used to generate the segmentation mask $M_{t}$.

	%%%%%% table performance
	\begin{table*}[htbp]
        \setlength{\tabcolsep}{4.1mm}
		\centering
		\begin{subtable}[t]{1.0\linewidth}
			\centering
			\begin{tabular}{l|c|cc|ccc|ccc}
				\toprule
				\multirow{2}{*}{Method}& \multirow{2}{*}{FB}    & \multirow{2}{*}{FPS} & \multirow{2}{*}{Mem}     & \multicolumn{3}{c|}{Before}                                    & \multicolumn{3}{c}{Finetune}                                     \\ 
				\cline{5-10}
				&  & &  & $\mathcal{J \& F} \uparrow$  &   $\mathcal{J} \uparrow$  & $\mathcal{F} \uparrow$            & $\mathcal{J \& F} \uparrow$  &    $\mathcal{J} \uparrow$  & $\mathcal{F} \uparrow$         \\ 
				\midrule
				LWL~\cite{bhat2020learning}  & \textbf{OD}  & 14.1   & 1.2   & 54.1        & 49.6        & 58.6            & 56.4       & 51.8         & 60.9          \\
				CFBI~\cite{yang2020collaborative} & \textbf{F+P}     & 5.2   & 3.82   & 50.0       & 45.0        & 55.1                 & 51.5      & 46.2         &  56.7       \\
				\iffalse
				AOT-T~\cite{yang2021associating} & \textbf{F+P}   & 53.2   & 0.85   & 51.5        & 46.2        & 56.8               & 53.6       & 48.2         & 59.1          \\
				\fi
				AOT-B~\cite{yang2021associating}  & \textbf{F+P}   & 31.9   & 0.96   & 56.9        & 51.8        & 61.9          & 58.9       & 53.5         & 64.2         \\
				AOT-L~\cite{yang2021associating}  & \textbf{A}   & 20.8   & 1.32   & 59.4        & 53.6        & 65.2           & 60.9       & 55.1         & 66.8        \\
				STCN~\cite{cheng2021rethinking} & \textbf{A}   & 22.1   & 0.92   & 45.8        & 41.1        & 50.5            & 48.9       & 43.9         & 54.0      \\
				\multicolumn{1}{l|}{AFB-URR~\cite{liang2020video}} & \textbf{C}   &  4.8  & 2.89    & 34.8        & 31.3        & 38.2            & 36.2       & 33.1         & 39.3      \\
				RDE~\cite{li2022recurrent}  & \textbf{C}  & 22.2   & 1.0   & 52.9        & 47.7        & 58.1         & 53.7       & 48.3         & 59.2        \\
				XMem~\cite{cheng2022xmem} & \textbf{C}  & 28.6   & 1.4   & 50.0        & 45.5        & 54.4        & 52.9       & 48.1         & 57.7       \\ 
				\midrule
				DDMemory  & \textbf{C} & \textbf{30.3}   & \textbf{0.88}   & \textbf{60.7}             & \textbf{55.0}        & \textbf{66.3}          & \textbf{61.9}       & \textbf{56.3}         & \textbf{67.4}   \\ \bottomrule
			\end{tabular}
			\caption{Results on validation set.}
			\label{tab:val}
		\end{subtable}
		\begin{subtable}[t]{1.0\linewidth}
			\centering
			\begin{tabular}{l|c|cc|ccc|ccc}
				\toprule
				\multirow{2}{*}{Method} & \multirow{2}{*}{FB} & \multirow{2}{*}{FPS}    &  \multirow{2}{*}{Mem}     & \multicolumn{3}{c|}{Before}                                    & \multicolumn{3}{c}{Finetune}                                     \\ 
				\cline{5-10}
				&  & &  & $\mathcal{J \& F} \uparrow$  &   $\mathcal{J} \uparrow$  & $\mathcal{F} \uparrow$          & $\mathcal{J \& F} \uparrow$  &    $\mathcal{J} \uparrow$  & $\mathcal{F} \uparrow$        \\ 
				\midrule
				\multicolumn{1}{l|}{LWL~\cite{bhat2020learning}} & \textbf{OD}   & 14.1   & 1.2   & 50.7       & 46.5        & 54.8             & 50.8       & 46.4         & 55.2       \\
				\multicolumn{1}{l|}{CFBI~\cite{yang2020collaborative}} & \textbf{F+P}   & 5.2   & 3.82   & 44.8        & 40.5        & 49.0                 & 44.8       & 40.2         &  49.4      \\
				\iffalse
				\multicolumn{1}{l|}{AOT-T~\cite{yang2021associating}} & \textbf{F+P}   & 53.2   & 0.85   & 49.8        & 45.2        & 54.4             & 50.2       & 45.4         & 55.0       \\
				\fi
				\multicolumn{1}{l|}{AOT-B~\cite{yang2021associating}} & \textbf{F+P}   & 31.9   & 0.96   & 54.4        & 49.3        & 59.4               & 54.5       & 49.2         & 59.8       \\
				\multicolumn{1}{l|}{AOT-L~\cite{yang2021associating}}  & \textbf{A}  & 20.8   & 1.32   & 54.1        & 48.7        & 59.5           & 54.7       & 49.2         & 60.2        \\
				\multicolumn{1}{l|}{STCN~\cite{cheng2021rethinking}} & \textbf{A}  & 22.1   & 0.92   & 45.8        & 41.6        & 50.0            & 48.3       & 44.0        & 52.5    \\
				\multicolumn{1}{l|}{AFB-URR~\cite{liang2020video}} & \textbf{C}   & 4.8   & 2.89   & 39.9        & 36.2        & 43.6            & 40.8       & 37.5         & 44.1    \\
				\multicolumn{1}{l|}{RDE~\cite{li2022recurrent}} & \textbf{C}   & 22.2   & 1.0   & 49.0        & 44.4        & 53.5      & 50.2       & 45.7         & 54.6    \\
				\multicolumn{1}{l|}{XMem~\cite{cheng2022xmem}} & \textbf{C}  & 28.6   & 1.4   & 49.5        & 45.2        & 53.7         & 50.9       & 46.5         & 55.3       \\ 
				\midrule
				\multicolumn{1}{l|}{DDMemory} & \textbf{C}  & \textbf{30.3}   & \textbf{0.88}   & \textbf{55.0}              & \textbf{49.9}        & \textbf{60.2}         & \textbf{55.7}     & \textbf{50.3}        & \textbf{61.2}        \\ \bottomrule
			\end{tabular}
			\caption{Results on test set.}
			\label{tab:test}
		\end{subtable}
		\caption{Comparisons with state-of-the art models on LVOS validation and test sets. FB denotes the kind of feature bank. \textbf{OD}, \textbf{F+P}, \textbf{A}, and \textbf{C} denote online adaption, first and previous frame, all frames, and  compressed memory bank respectively. We re-time these models on our hardware (a V100 GPU) for a fair comparison. Mem denotes the maximum GPU memory usage (in GB). Before and Finetune denotes the results just trained on short-term video datasets and finetuned on LVOS training set.}
		\label{tab:performance2}
	\end{table*}

	\subsection{Global Memory Encoder}
	\label{sec:longencoder}
	To use fixed-size features to construct global historical information with as little loss as possible, we adopt a recurrent manner to build a global feature memory bank. Specifically, at time $t$, we utilize a Gated Recurrent Unit (GRU) \cite{chung2014empirical,shi2015convolutional} as global memory encoder to compress $\{{f_{M}^{i}}\}^{t-2}_{i=1}$ into global memory $Mem_{G}^{t}$.
	
	For the generation and updating of $Mem_{G}^{t}$, we first initialize 
	$Mem_{G}^{1}$ as $f_{M}^{1}$, and then utilize GRU to propagate it as illustrated in Figure \ref{fig:model}. The process is defined as:
	\begin{equation}
		\setlength\abovedisplayskip{0.2cm}
		\setlength\belowdisplayskip{0.2cm}
		Mem_{G}^{t} = GRU(Mem_{G}^{t-1},f_{M}^{t-2})
		\label{eq:gru}
	\end{equation}
	where $GRU$ denotes a GRU module. By recurrent refreshment, we achieve the ability to encode global information into a  fixed size features and discard redundant and noisy information.

    \subsection{Diverse Dynamic Memory}
      
	Our diverse dynamic memory consists of three types of different temporal scale memory banks: \textit{reference memory}, \textit{global memory}, and \textit{local memory}. Due to the fixed-size memory features, the memory cost remains constant no matter how long the video is. The first frame and its groundtruth mask are stored in reference memory, which is responsible for the  recovery after disappearance or occlusion. Global memory leverages a recurrent manner to store historical information effectively, which is crucial for the segmentation of long-term videos. Local memory is updated every frame and provides the location and shape cues. The complementary memory banks achieve the storage  of rich temporal information and the removal of noisy features. Thanks to the diverse and dynamic memory banks, DDMemory achieves promising performance with a constant memory cost and high speed in long videos.

    \subsection{Model Details} We employ MobileNet-V2~\cite{sandler2018mobilenetv2} as backbone. Memory encoder and query encoder share the same weight, following \cite{yang2021associating}. LSTT module proposed in \cite{yang2021associating} is adopted as matching module, and the LSTT layer number is 3. We use FPN~\cite{lin2017feature} as decoder. %which is made up of a sequence of residual convolutional blocks and each block gradually upsamples feature by a factor of two. 
    For Global Memory Encoder, when segmenting current frame $I_{t}$, we just save the latest global memory $Mem^{t}_{G}$ for the sake of efficiency. After the segmentation of $I_{t}$, we update $Mem^{t}_{G}$ by utilizing the GRU module to obtain the global memory $Mem^{t+1}_{G}$ for next frame $I_{t+1}$. We don’t need to restore all intermediate frames and repeat the calculation of global memory. We only need to store a fixed-size global memory and conduct a simple updating per frame.

	%experiment
	\section{Experiments}
	\subsection{Experiment Setup}
	\textbf{Experiment Settings.}
	We evaluate our DDMemory and other existing VOS methods, including CFBI~\cite{yang2020collaborative}, LWL \cite{bhat2020learning}, STCN \cite{cheng2021rethinking}, AOT \cite{yang2021associating}, RDE \cite{li2022recurrent}, XMem \cite{cheng2022xmem} on LVOS validation and test set. 
	We restrict the memory length to 6 when evaluating approaches with memory bank, such as STCN~\cite{cheng2021rethinking}, AOT \cite{yang2021associating}, and XMem~\cite{cheng2022xmem}. For a fair comparison, all the videos are down-sampled to 480p resolution. we also finetune these models on the training set of LVOS for two epochs, with a learning rate of  $5 \times 10^{-4}$, and reevaluate their performance.

	\textbf{Evaluation Metrics.} We adopt the two commonly used evaluation metrics, region similarity $\mathcal{J}$ and contour accuracy $\mathcal{F}$ as metrics, following DAVIS~\cite{perazzi2016benchmark,perazzi2017learning} and YouTube-VOS~\cite{xu2018youtube}. And we calculate the average their mean value as the final score.

	%%%% qualitative result
	\begin{figure*}[ht]
		\centering
		\includegraphics[width=\linewidth]{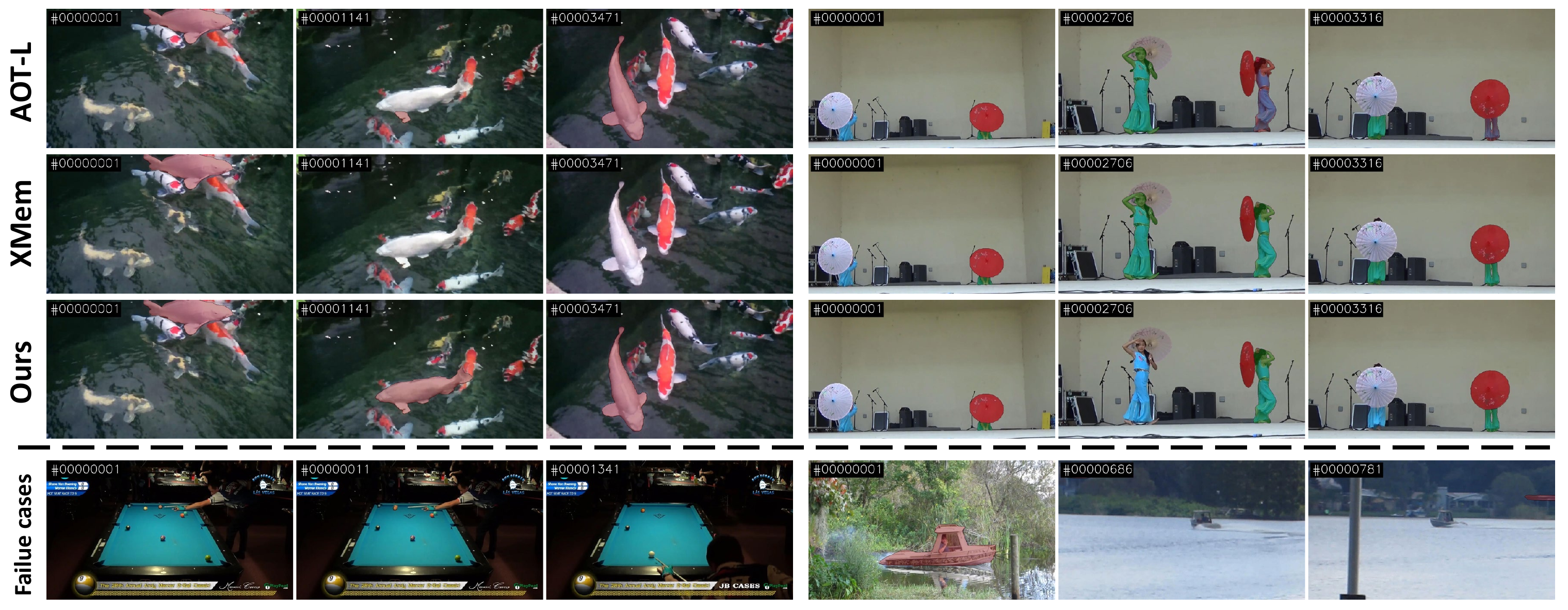}
		\caption{Qualitative results on LVOS validation and test set. We compare the result of DDMemory with XMem\cite{cheng2022xmem} and AOT-L\cite{yang2021associating} in top section, and DDMemory performs better. The bottom row shows failure cases.}
		\label{fig:show}
	\end{figure*}

\subsection{Benchmark Results}
	\textbf{Quantitative Results.} 
	As shown in Table \ref{tab:performance2}, DDMemory outperforms other models with different feature banks on both validation (60.7 $\mathcal{J} \& \mathcal{F}$) and test sets (55.0 $\mathcal{J} \& \mathcal{F}$) when maintaining a real-time speed (30.3 FPS) at the lowest GPU memory cost (0.88G). 
	Thanks to the diverse and dynamic memory banks, DDMemory exploits richer temporal contexts and are robust to complex challenges in long-term videos. We finetune these models and DDMemory on the LVOS training sets and assess their performances again. Although each model's performance has essentially improved to some degree, DDMemory continues to outperform all the competitors on both validation (61.9 $\mathcal{J} \& \mathcal{F}$) and test (55.7 $\mathcal{J} \& \mathcal{F}$) sets.

\textbf{Qualitative Results.}
	We present the segmentation results in comparison with AOT-L and XMem in Figure \ref{fig:show}. As shown in the top section, AOT and XMem lose track or confuse similar objects, while DDMemory can re-detect the target after out-of-view and handle multiple similar objects successfully. The bottom section displays some failure cases. Although DDMemory achieves promising performance, DDMemory is still not robust enough for the complex motion. More details are in supplementary material.

	%%%%%%%% ablation
	\begin{table}[htbp]
        \setlength{\tabcolsep}{3.5mm} 		
		\tiny
		\centering
		\begin{tabular}{ccccccccc}
			\toprule
			R & G & L & FPS & GPU & $\mathcal{J \& F}$ & $\mathcal{J}$ & $\mathcal{F}$  \\ 
			\midrule
			\CheckmarkBold & ~ & ~ & 57.4 & 0.52 & 44.2 & 39.0 & 49.4   \\ 
			~ & \CheckmarkBold & ~ & 55.2 & 0.62 & 42.7 & 37.4 & 48.0   \\ 
			~ & ~ & \CheckmarkBold & 43.5 & 0.68 & 18.3 & 17.1 & 19.6   \\ 
			\CheckmarkBold & \CheckmarkBold & ~ & 46.7 & 0.78 & 47.8 & 42.4 & 53.3   \\ 
			\CheckmarkBold & ~ & \CheckmarkBold & 35.6 & 0.82 & 57.9 & 53.0 & 62.8   \\ 
			~ & \CheckmarkBold & \CheckmarkBold & 35.1 & 0.76 & 54.9 & 51.1 & 58.7   \\ 
			\midrule
			\CheckmarkBold & \CheckmarkBold & \CheckmarkBold & 30.3 & 0.88 & \textbf{61.9} & \textbf{56.3} & \textbf{67.4}  \\ \bottomrule
		\end{tabular}
		\caption{Ablation study on LVOS validation set. R, G, and L denote $Mem_{R}$, $Mem_{G}$, and $Mem_{L}$, respectively.}
		\label{tab:ablation}
	\end{table}

	\textbf{Ablation Study.}
	We perform an ablation study on LVOS validation set and analyze the contribution of each memory bank to the segmentation result in Table \ref{tab:ablation}. We experiment with various combinations of three memory banks. Results show the role of each component. The reference memory $Mem_{R}$ is responsible for the re-detection after occlusion or out-of-view and is sensitive to large appearance changes. The global memory $Mem_{G}$ encodes the long-term temporal information. The local memory $Mem_{L}$ provides location cues and appearance prior. For long-term VOS, all three memory banks are crucial and complementary. Please see supplementary material for more details.

	%%%%%%%% oracle
	\begin{table}[htbp]
        \setlength{\tabcolsep}{2.8mm} 
		%\small
		\centering
		\begin{tabular}{cccccc}
			\toprule
			Oracle Box & Oracle Mask & $\mathcal{J \& F}$ & $\mathcal{J}$ & $\mathcal{F}$  \\ \midrule
			~  &  ~ & 61.9 & 56.3 & 67.4 \\
			\CheckmarkBold & ~ & 70.2 & 64.6 & 75.7  \\ 
			~ & \CheckmarkBold & 82.7 & 76.5 & 89.0  \\
			\CheckmarkBold & \CheckmarkBold & 84.4 & 77.8 & 91.1  \\ \bottomrule
		\end{tabular}	
		\caption{Oracle analysis on LVOS validation set.}
		\label{tab:oracle}
	\end{table}

	\textbf{Oracle Analysis.}
	To conduct further analysis of object localization and association, we carry out oracle experiments. Results are shown in Table \ref{tab:oracle}. Average performance is improved by 8.3 \% when segmentation is provided with an oracle bounding box, proving that segmentation errors result from poor tracking between similar objects. While resolving the segmentation errors, the model achieves a higher score (20.8 \% boost). This shows that error accumulation is the primary cause of errors. But even if the correct masks and locations are provided, there is still a large gap between the result (84.4 $\mathcal{J} \& \mathcal{F}$) and groundtruth. The gap demonstrates that complex movements are still very challenging for VOS models. In short, error accumulation is the main cause of unsatisfactory performances, and a robust VOS model must be able to handle the much more complex motion in long-term videos.

	\subsection{Results on DAVIS Short-term Validation Set}
	Experiments in Table \ref{tab:performance2} show the effectiveness of our algorithm on long-term videos. To show the efficacy of DDMemory on short-term videos, we evaluate DDMemory and other models on DAVIS 2017 validation set \cite{perazzi2017learning}. 
	%DAVIS 2017 is a popular short-term VOS dataset, which has 30 videos in validation split. 
	The result is shown in Table \ref{tab:davis}. DDMemory exceeds the majority of models and maintains an efficient speed (28.1 FPS). Despite having higher performance than DDMemory, XMem and STCN employ a stronger  backbone ResNet50~\cite{he2016deep}, while DDMemory only uses MobileNet-V2~\cite{sandler2018mobilenetv2}. 
	%The reason that the improvement on DAVIS 2017 validation set resulting from the global temporal information is not very obvious. The reason may be that the length of videos is relatively short. More experiment results are in supplementary material.
	The reason why the improvement resulting from the global temporal information is not very obvious may be that the length of videos is relatively short. More experiment results are in supplementary material.

    %%%%%%%% davis
	\begin{table}[htbp]
        \footnotesize
		\centering
		\begin{tabular}{lccccc}
			\toprule
			Methods & Backbone &  $\mathcal{J \& F}$   &  $\mathcal{J}$   & $\mathcal{F}$    & FPS  \\ \midrule
			CFBI~\cite{yang2020collaborative} & ResNet101~\cite{he2016deep}   & 81.9 & 79.1 & 84.6 & 5.9  \\
			LWL~\cite{bhat2020learning}   & ResNet50~\cite{he2016deep}   &  81.6 & 79.1 & 84.1 & 13.2 \\
			STCN~\cite{cheng2021rethinking} & ResNet50~\cite{he2016deep}   & \underline{85.4} & \underline{82.2} & \underline{88.6} & 20.2 \\
			RDE~\cite{li2022recurrent} & ResNet50~\cite{he2016deep}    & 84.2 & 80.8 & 87.5 & 27.0   \\
			XMem~\cite{cheng2022xmem}  & ResNet50~\cite{he2016deep}  & \textbf{86.2} & \textbf{82.9} & \textbf{89.5} & 22.6 \\
			AOT-B~\cite{yang2021associating}  & MobileNet-V2~\cite{sandler2018mobilenetv2}  & 82.5 & 79.7 & 85.2 & \textbf{29.6} \\
			AOT-L~\cite{yang2021associating} & MobileNet-V2~\cite{sandler2018mobilenetv2}   & 83.8 & 81.1 & 86.4 & 18.7 \\
			%STM     & 81.8 & 79.2 & 84.3 & 11.1 \\
			%HMMN    & 84.7 & 81.9 & 87.5 & 9.3  \\
			%RMNet   & 83.5 & 81.0 & 86.0 & 4.4  \\
			%JOINT   & 83.5 & 80.8 & 86.2 & 6.8  \\ 
			\midrule
			DDMemory & MobileNet-V2~\cite{sandler2018mobilenetv2}   & 84.2 & 81.3 & 87.1 & \underline{28.1}  \\ \bottomrule
		\end{tabular}
		\caption{Comparisons with state-of-the art models on DAVIS 2017 validation set. Bold and underline denote the best and second-best respectively in each column.}
		\label{tab:davis}
	\end{table}

    \subsection{Attribute-based Evaluation} 
    We report the performance in Table~\ref{tab:attribute} on the validation set characterized by with the most informative atttributes. FM (Fast Motion), OCC (Occlusion), OV (Out-of-view), SV (Scale Variation), and AC (Appearance change) are all well known challenges in short-term video segmentation, which also have great influence on performance of long-term task. Furthermore, long-term videos present specific challenges, such as Long-term Reappearance (LRA) and Cross-temporal Confusion (CTC), which have a worse impact. Although these methods achieve promising results on short-term VOS datasets (over 80\% $\mathcal{J} \& \mathcal{F}$),  they still struggle with complex scenes and frequent reappearance in long-term videos, highlighting the unique value of our LVOS. The ability to recover disappeared object, distinguish target from similar background, detect small object, and model long-term historical information is crucial for robust LVOS.

	%%% attribute
\begin{table}
    \setlength{\tabcolsep}{1.8mm} 
	\centering
	\scriptsize
	%\vspace{-2.8em} 
	\begin{tabular}{cccccccc}
		\toprule
		Attr & AOT-B & AOT-L & XMem & LWL & Ora B & Ora M & Ora B+M\\
		\midrule
		FM & 54.3 & 55.3 & 46.7 & 48.2 & 73.8 & 82.6 & 85.5\\
		OCC & 50.6 & 52.1 & 47.6 & 50.3 & 72.5 &79.1 & 83.6\\
		OV & 54.4 & 55.2 & 53.8 & 48.6 & 74.2 & 79.9 & 82.8\\
		SV &48.3 & 50.4 & 44.7 & 47.2 & 66.8 & 76.6 & 80.5\\
		AC & 53.1 & 55.7 & 48.9 & 52.4 & 75.7 & 82.2 & 84.2\\
		LRA & 44.3 & 45.3 & 40.7 & 45.2 & 63.8 & 74.6 & 78.5\\
		CTC & 44.5 & 45.7 & 45.1 & 46.1 & 64.4 & 75.5 & 77.7\\
		\bottomrule        
	\end{tabular}
	\caption{Attribute-based aggregate performance. For each method, we just show $\mathcal{J}$. Ora B, Ora M, Ora B+M denote oracle box, oracle mask and oracle box + mask in oracle experiments (Table~\ref{tab:oracle}), respectively.}
	\label{tab:attribute} 
\end{table}

	\section{Conclusion}
	In this paper, we propose a new long-term video object segmentation dataset, LVOS. Different from existing short-term VOS datasets, the average length of videos in LVOS is  1.59 minutes. More complex motion and longer duration place greater demands on VOS models. We assess existing VOS approaches and propose a novel baseline method DDMemory designed for long-term VOS. Based on the baseline model, we analyze the weakness of prior methods and point promising directions for further study. We hope that LVOS can provide a platform to encourage a comprehensive study on long-term VOS.

\section{Acknowledge}
This work was supported by National Natural Science Foundation of China (No.62072112), Scientific and Technological Innovation Action Plan of Shanghai Science and Technology Committee (No.22511102202), National Key R\&D Program of China (2020AAA0108301).

{\small
\bibliographystyle{ieee_fullname}
\bibliography{egbib}
}

%% appendix
   \onecolumn
   \begin{appendices}
   
\section{Dataset Construction and Annotations}
    For the class selection, we summarize the classes of LaSOT~\cite{fan2019lasot} and VOT-LT 2019~\cite{kristan2019seventh}. There are 85 classes in LaSOT and about 20 classes in VOT-LT 2019. Then we carefully select a set of categories based on the following rules:  (1) The resolution of videos are larger than 720p, (2) The video is representative enough to include at least one attribute demonstrated in Table 2 of paper. (3) Class of the video is relative to daily life. (4) The total number of videos with this category should be greater than ten. Based on the four rules, we choose 27 categories. Because VOT-LT and LaSOT are single-object tracking dataset, LVOS is a mutliple-object. For target selection, we may follow the target object in VOT-LT and LaSOT, or select different objects as targets.
    
    For the annotation process, because all the masks are obtained by models, we need two-pass manual corrections. During Step 1 1 FPS automatic sgmentation, we utilize the box of target object in each frame to get segmentation. If the target object of a video is the same as that in LaSOT or VOT-LT, we use the gropundtruth boxes. Otherwise, we adopt tracking model to obtain the box of target object in each frame.

\section{Training Strategy}
	Following \cite{yang2021associating,li2022recurrent}, we divide the training stage into two phases: (1) pretraining on static image datasets~\cite{cheng2014global,everingham2010pascal,lin2014microsoft,shi2015hierarchical,hariharan2011semantic} by applying data augmentation such as synthetic deformation with the initial learning rate of $4 \times 10^{-4}$ and a weight decay of 0.03 for 100,100 steps. (2) main training on the VOS datasets \cite{perazzi2017learning,xu2018youtube} with the initial learning rate of $2 \times 10^{-4}$ and a weight decay of 0.07 for 100,100 steps.  AdamW~\cite{loshchilov2017decoupled} optimizer is adopted for optimization. The batch size is set as 16. Dice loss \cite{nowozin2014optimal} and bootstrapped cross entropy loss with equal weighting is used.

%%%%%%%% davis
	\begin{table}[htbp]
		\centering
		\begin{tabular}{lccccc}
			\toprule
			Methods & Backbone &  $\mathcal{J \& F}$   &  $\mathcal{J}$   & $\mathcal{F}$    & FPS  \\ \midrule
			CFBI\cite{yang2020collaborative} & ResNet101\cite{he2016deep}   & 81.9 & 79.1 & 84.6 & 5.9  \\
			LWL\cite{bhat2020learning}   & ResNet50\cite{he2016deep}   &  81.6 & 79.1 & 84.1 & 13.2 \\
			STCN\cite{cheng2021rethinking} & ResNet50\cite{he2016deep}   & \underline{85.4} & \underline{82.2} & \underline{88.6} & 20.2 \\
			RDE\cite{li2022recurrent} & ResNet50\cite{he2016deep}    & 84.2 & 80.8 & 87.5 & 27.0   \\
			XMem\cite{cheng2022xmem}  & ResNet50\cite{he2016deep}  & \textbf{86.2} & \textbf{82.9} & \textbf{89.5} & 22.6 \\
			AOT-B\cite{yang2021associating}  & MobileNet-V2\cite{sandler2018mobilenetv2}  & 82.5 & 79.7 & 85.2 & \textbf{29.6} \\
			AOT-L\cite{yang2021associating} & MobileNet-V2\cite{sandler2018mobilenetv2}   & 83.8 & 81.1 & 86.4 & 18.7 \\
			%STM     & 81.8 & 79.2 & 84.3 & 11.1 \\
			%HMMN    & 84.7 & 81.9 & 87.5 & 9.3  \\
			%RMNet   & 83.5 & 81.0 & 86.0 & 4.4  \\
			%JOINT   & 83.5 & 80.8 & 86.2 & 6.8  \\ 
			\midrule
			DDMemory & MobileNet-V2\cite{sandler2018mobilenetv2}   & 84.2 & 81.3 & 87.1 & \underline{28.1}  \\ \bottomrule
		\end{tabular}
		\caption{Comparisons with state-of-the art models on DAVIS 2017 validation set\cite{perazzi2017learning}. Bold and underline denote the best and second-best respectively in each column.}
		\label{tab:davis}
	\end{table}

%%%%%%%% ytb
	\begin{table}[htbp]
		
		\centering
		\begin{tabular}{lccccccc}
			\toprule
			Methods & Backbone &  $\mathcal{J \& F}$   &  $\mathcal{J}_{s}$   & $\mathcal{F}_{s}$    & $\mathcal{J}_{u}$   & $\mathcal{F}_{u}$   & FPS  \\ \midrule
			CFBI\cite{yang2020collaborative} & ResNet101\cite{he2016deep}   & 81.4 & 81.1 & 85.8 &  75.3  &  83.4  & 4.0  \\
			LWL\cite{bhat2020learning}   & ResNet50\cite{he2016deep}   &  81.5 & 80.4 & 84.9  & 76.4 & 84.4 & - \\
			STCN\cite{cheng2021rethinking} & ResNet50\cite{he2016deep}   & 83.0 & 81.9 & 86.5 & 77.9  &  85.7  & 13.2 \\
			RDE\cite{li2022recurrent} & ResNet50\cite{he2016deep}    & 81.9 & 81.1 & 85.5 &  84,8  & 76.2 &  17.7   \\
			XMem\cite{cheng2022xmem}  & ResNet50\cite{he2016deep}  & \textbf{85.7} & \textbf{84.6} & \textbf{89.3} & \textbf{80.2} &  \textbf{88.7} & 11.8 \\
			AOT-B\cite{yang2021associating}  & MobileNet-V2\cite{sandler2018mobilenetv2}  & 83.5 & 82.6 & 87.5 & 77.7 & 86.0 & \textbf{20.5} \\
			AOT-L\cite{yang2021associating} & MobileNet-V2\cite{sandler2018mobilenetv2}   & 83.8 & 82.9 & 87.9 & 77.7  &  86.5 &  16.0 \\
			%STM     & 81.8 & 79.2 & 84.3 & 11.1 \\
			%HMMN    & 84.7 & 81.9 & 87.5 & 9.3  \\
			%RMNet   & 83.5 & 81.0 & 86.0 & 4.4  \\
			%JOINT   & 83.5 & 80.8 & 86.2 & 6.8  \\ 
			\midrule
			DDMemory & MobileNet-V2\cite{sandler2018mobilenetv2}   & \underline{84.1} & \underline{83.5} & \underline{88.4} & \underline{78.1} & \underline{86.5} & \underline{18.7}  \\ \bottomrule
		\end{tabular}
		\caption{Comparisons with state-of-the art models on YouTubeVOS-2018 validation set \cite{xu2018youtube}. Bold and underline denote the best and second-best respectively in each column.}
		\label{tab:ytb}
		
	\end{table}
	
	\section{Results on Short-term Videos Validation Sets}
	We compare our DDMemory with state-of-the-art VOS models on short-term videos validation datasets (DAVIS 2017~\cite{perazzi2017learning} and YouTube-VOS 2018~\cite{xu2018youtube}) in Table \ref{tab:davis} and \ref{tab:ytb}. We re-time these models on our hardware (one V100 GPU) for a fair comparison. DDMemory exceeds the majority of models and maintains an efficient speed. Despite having higher performance than DDMemory, XMem and STCN employ a stronger  backbone ResNet50~\cite{he2016deep}, while DDMemory only uses MobileNet-V2~\cite{sandler2018mobilenetv2}. Although the segmentation accuracy in short-term videos can be improved by the global temporal information, but the improvement on short-term videos validation sets is not very obvious. The reason may be that the length of the videos is	relatively short.

	%%%% qualitative result
	\begin{figure*}[htbp]
		\centering
		\includegraphics[width=1.0\linewidth]{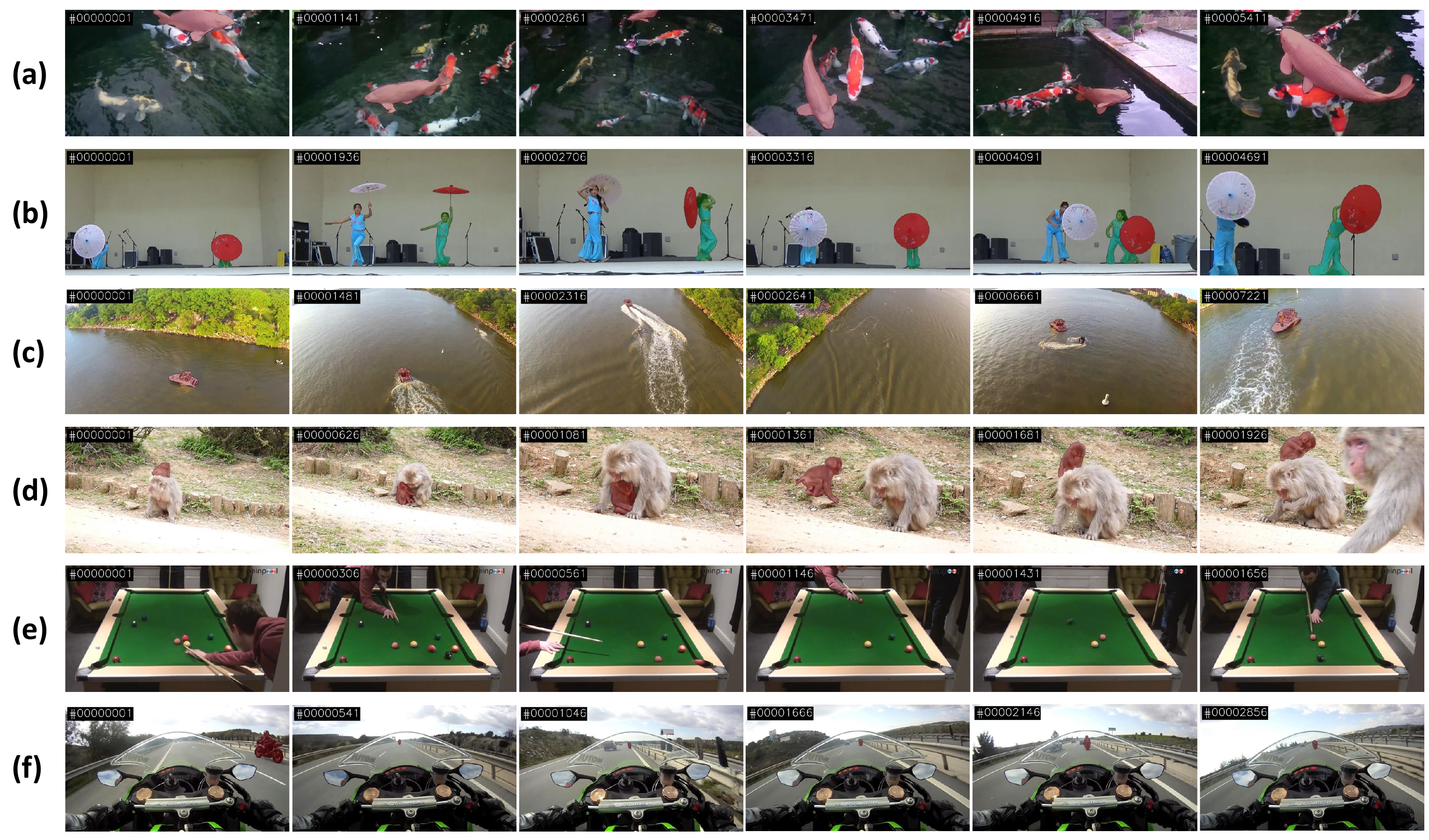}
		
		\caption{Qualitative results on LVOS validation and test set. DDMemory performs well on long-term videos. }
		\label{fig:show}
	\end{figure*}

	\section{Additional Qualitative Results}
	We show more qualitative results on LVOS in Figure \ref{fig:show}. As demonstrated, our DDMemory can handle many challenging long-term VOS attributes, such as long-term reappearance, similar objects, occlusion, fast and complex occlusions, low resolution, and scale variation, etc. In row (a), DDMemory successfully distinguishes the white goldfish with other similar fishes. In row (b), the two people and umbrellas are not confused with each other in spite of occlusion. In row (c), DDMemory can re-detect the boat after long-term and frequent disappearance. In row (d),  the small white ball is  similar to other balls, and DDMemory still succeeds in tracking and segmenting it. In row (f), DDMemory tracks the motorcycle well despite the fast motion and large scale variation.

%%%%%%%% ablation
	\begin{table}[htbp]
		\centering
		\begin{tabular}{cccccccc}
			\toprule
			R & G & L & FPS & GPU & $\mathcal{J \& F}$ & $\mathcal{J}$ & $\mathcal{F}$  \\ 
			\midrule
			\CheckmarkBold & ~ & ~ & 57.4 & 0.52 & 44.2 & 39.0 & 49.4  \\ 
			~ & \CheckmarkBold & ~ & 55.2 & 0.62 & 42.7 & 37.4 & 48.0  \\ 
			~ & ~ & \CheckmarkBold & 43.5 & 0.68 & 18.3 & 17.1 & 19.6  \\ 
			\CheckmarkBold & \CheckmarkBold & ~ & 46.7 & 0.78 & 47.8 & 42.4 & 53.3  \\ 
			\CheckmarkBold & ~ & \CheckmarkBold & 35.6 & 0.82 & 57.9 & 53.0 & 62.8  \\ 
			~ & \CheckmarkBold & \CheckmarkBold & 35.1 & 0.76 & 54.9 & 51.1 & 58.7  \\ 
			\CheckmarkBold & \CheckmarkBold & \CheckmarkBold & 30.3 & 0.88 & 61.9 & 56.3 & 67.4 \\ \bottomrule
		\end{tabular}
		\caption{Ablation study on LVOS validation set. R, G, and L denote $Mem_{R}$, $Mem_{G}$, and $Mem_{L}$, respectively.}
		\label{tab:ablation}
	\end{table}

%%%% ablation picture
	\begin{figure*}[htbp]
		\centering
		\includegraphics[width=1.0\linewidth]{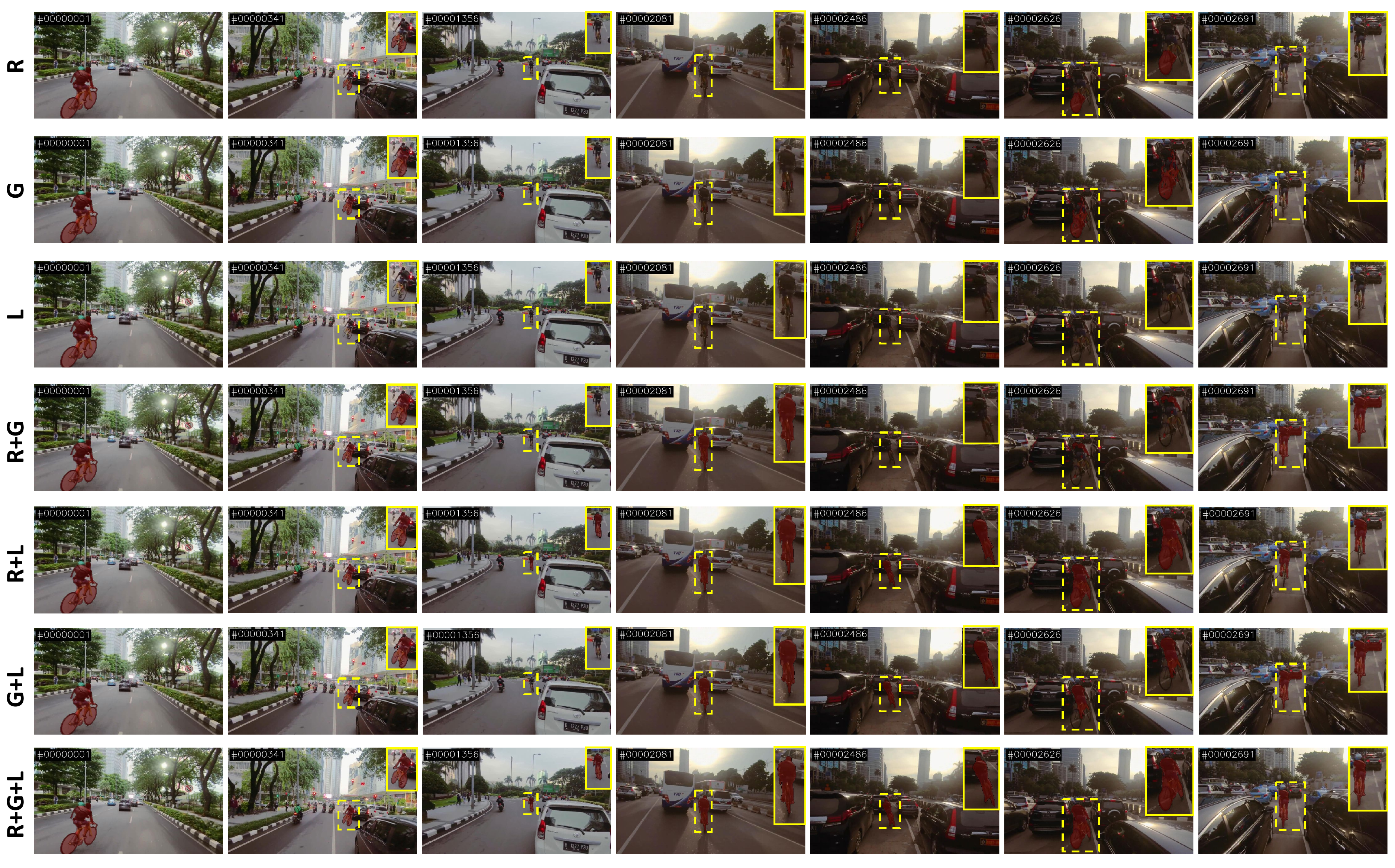}
		
		\caption{Ablation study. We visualize the results of different combinations of three memory banks on the same video. Best viewed in color. }
		\label{fig:ablation}
	\end{figure*}

	\section{More Analysis About Diverse Dynamic Memory}

	We conduct an ablation study on the role of each memory bank in Table \ref{tab:ablation}. To more clearly illustrate each memory bank's impact, we visualize the results of different combinations of three memory banks on the same video in Figure \ref{fig:ablation}. This video is about a man riding a bike through the streets. The man frequently gets occluded by cars. Moreover, there are many other challenges in this video, such as background clutter (there are many similar people on the street), fast motion (the man is moving quickly), low resolution (sometimes the bounding-box of this man is small), and significant appearance change (the appearance of this man changes a lot over the time). This video is extremely challenging. In the first row, we just use the reference memory $Mem_{R}$, despite the re-detection after occlusion, the reference memory is sensitive to large appearance changes. In the second row, only global memory $Mem_{G}$ is enabled. Global memory is rich in temporal information so $Mem_{G}$ can handle occlusion, too. Because of the error accumulation, there are still many segmentation defects. In the third row, only local memory $Mem_{L}$ is used, and it is obvious that the model loses track after the first occlusion. In the fourth row, we utilize the reference and global memory. Although the model is better at handling changes in appearance, it still has trouble precisely segmenting the target. In the fifth row, we combine reference and local memory. The local memory boosts the contour accuracy to a large descent. In the sixth row, global memory and local memory banks are used. Compared to the fifth raw, the segmentation accuracy is a little worse. In the final row, we combine the three complementary memory banks. DDMemory tracks and segments target objects successfully. The visual results demonstrate the role of the three memory banks.  The reference memory $Mem_{R}$ is responsible for the re-detection after occlusion or out-of-view and is sensitive to large appearance changes. The local memory $Mem_{L}$ provides location cues and appearance prior. The global memory $Mem_{G}$ encodes the long-term temporal information as a complement to the other two memory features. For long-term VOS, all three memory banks are essential and complementary.

\section{Oracle Experiments}
    For oracle box, we convert groundtruth mask into box and only search target in the groundtruth box area. For oracle mask, we search target in whole image and use groundtruth mask to update $Mem_{G}$ and $Mem_{L}$. For oracle box and mask, we search target in the groundtruth box area and use groundtruth mask to update $Mem_{G}$ and $Mem_{L}$

\section{Attribute-based Evaluation}
    We report performance of more models in Table~\ref{tab:attribute} on validation set characterized by the most informative attributes. Scale variation has a more pronounced negative impact on short-term visual object segmentation (VOS) performance than other challenges, particularly for models that employ online adaption (OD) or compressed memory (C) feature banks. Additionally, specific long-term challenges have an even greater impact on accuracy. Visual object segmentation (VOS) models may lose track of the target object when it becomes small in size. Models that always keep the first frame in memory can re-detect the target object. However, models that employ online adaption (OD) or compressed memory (C) feature banks may mistake background objects for the target object, or they may be unable to restore detection due to the lack of guidance from the first frame. Therefore, the ability to recover a disappeared object, distinguish the target object from similar background objects, detect small objects, and model long-term historical information is crucial for robust LVOS.
    
%%% attribute
\begin{table}
    \setlength{\tabcolsep}{1.67mm} 
	\centering
    \footnotesize
	%\vspace{-2.8em} 
	\begin{tabular}{cccccccccccc}
		\toprule
		Attr & AFB-URR~\cite{liang2020video} & RDE~\cite{li2022recurrent} & CFBI~\cite{yang2020collaborative} & AOT-B~\cite{yang2021associating} & AOT-L~\cite{yang2021associating}  &  STCN~\cite{cheng2021rethinking} & XMem~\cite{cheng2022xmem} & LWL~\cite{bhat2020learning} & Ora B & Ora M & Ora B+M\\
		\midrule
		FM & 34.1 & 48.4 & 45.3 &  54.3 & 55.3 & 42.5 & 46.7 & 48.2 & 73.8 & 82.6 & 85.5\\
		OCC & 34.5 & 48.2 & 46.1 & 50.6 & 52.1 & 43.3 & 47.6 & 50.3 & 72.5 &79.1 & 83.6\\
		OV & 42.2 & 53.4 & 47.6 & 54.4 & 55.2 & 51.5 & 53.8 & 48.6 & 74.2 & 79.9 & 82.8\\
		SV & 33.1 & 48.4 & 45.7 & 48.3 & 50.4 & 41.5 & 43.7 & 47.2 & 66.8 & 76.6 & 80.5\\
		AC & 41.6 & 51.9 & 45.9 & 53.1 & 55.7 & 48.2 & 48.9 & 52.4 & 75.7 & 82.2 & 84.2\\
		LRA & 33.1 & 41.4 & 39.9 & 44.3 & 45.3 & 37.5 & 40.7 & 45.2 & 63.8 & 74.6 & 78.5\\
		CTC & 36.9 & 41.6 & 40.4 & 44.5 & 45.7 & 39.7 & 45.1 & 46.1 & 64.4 & 75.5 & 77.7\\
		\bottomrule        
	\end{tabular}
	\caption{Attribute-based aggregate performance. For each method, we just show $\mathcal{J}$. Ora B, Ora M, Ora B+M denote oracle box, oracle mask and oracle box + mask in oracle experiments, respectively.}
	\label{tab:attribute}
\end{table}
   
   \end{appendices}

\end{document}